\documentclass[journal]{IEEEtran}

\usepackage{listings}

\usepackage{graphicx}
\usepackage{multirow} 
\usepackage{array}
\usepackage{overpic}

\usepackage{hyperref}
\hypersetup{hypertex=true,
            colorlinks=true,
            linkcolor=blue,
            anchorcolor=blue,
            citecolor=blue}
\usepackage{tcolorbox}

\usepackage{amsmath} 
\usepackage{amssymb}
\usepackage[T1]{fontenc}
\begin{document}

% >>:-----------------------------------------------------------------------------------
\title{Robot Detection System:  Front-Following1}

\author{Jinwei~Lin,~\IEEEmembership{Member,~IEEE,}
        
\thanks{Jinwei Lin is come from Monash University,} 
\thanks{ORCID of Jinwei Lin:0000 0003 0558 6699,} 
\thanks{Manuscript written in 2018.}} 

% Paper Title
\markboth{Journal or Conference of \LaTeX\, No.~1, May~2023}
{Shell \MakeLowercase{\textit{et al.}}: Simple  Arrow Area Architecture Template}
\maketitle

% <<:-----------------------------------------------------------------------------------

% >>:-----------------------------------------------------------------------------------
\begin{abstract}
Front-following is more technically difficult to implement than the other two human following technologies, but front-following technology is more practical and can be applied in more areas to solve more practical problems. Front-following technology has many advantages not found in back-following and side-by-side technologies. In this paper, we will discuss basic and significant principles and general design idea of this technology. Besides, various of novel and special useful methods will be presented and provided. We use enough beautiful figures to display our novel design idea. Our research result is open source in 2018, and this paper is just to expand the research result propagation granularity. Abundant magic design idea are included in this paper, more idea and analyzing can sear and see other paper naming with a start of Robot Design System with Jinwei Lin, the only author of this series papers.

\end{abstract}

% <<:-----------------------------------------------------------------------------------

% >>:-----------------------------------------------------------------------------------
% Note that keywords are not normally used for peerreview papers.

% <<:-----------------------------------------------------------------------------------

\section{Introduction}

Front-following technology can reduce the psychological burden of the followers and make the target person feel safer. There are two main aspects to security here. One is that if the users use robots of back-following technology, they will worry about whether the robots behind them will crash and hit them. On the other hand, if the robot adopts the front-following method, the user can supervise the robot in front of them that is performing the following tasks in the first perspective.Such a following mode is more useful when the user uses the robot to carry the load and needs to view the goods at any time. Front-following technology is a type of human-following technology. It is a robot technology and human-computer interaction technology with few researches and great research value and plenty of application scenarios. The human-following technique can be divided into three types: back-following, side-by-side and front-following,depending on the location of the robot when the robot follows humans. Side-by-side-and front-following are technically more difficult than back-following. One of the basic reasons is that when implementing side-by-side and front-following, the robot often needs to predict people's future actions, while back-following is basically unnecessary.

The research paper of this project describes the algorithm research and theoretical analysis of Front-following, an advanced human-computer interaction technology. In order to better reflect and describe the implementation principle of Front-following technology and the author's self-innovation technology, we adopt the combination of partial and whole, start from the basic algorithm, and then gradually increase the depth and difficulty of the explained algorithm. During the elaboration of the technical principles, various algorithms will call each other to implement the joint use of the algorithms. In this project paper, all of the algorithms are designed independently by the author. Since this project research paper was completed independently by the author, it's compose lasted for 36 days, what is more, it was not enough time to write the paper every day. Therefore, there are inevitable problems or insufficient in neglect or omission in this paper.We hope that you will give us advice and guidance if you find any shortcomings. The purpose of writing this project research paper is to share our research results and research ideas and strategies with you through this paper. We also hope that our paper can serve as a reference and auxiliary advice for relative researchers. Thank you very much for your reading! Thanks!

\section{Design Background}

This project aims to set up a human interaction robot based on front-following technology, which can realize the function of following the target person in front of the target person. The prototype of the project robot's product mechanical drive is positioned in a simple front-following four-wheeled robot. Design and build a physical robot. The robot has a front-following function. It has the function of predicting the path and direction of the target person. The robot can use sensors, but not RGBD cameras or other types of cameras. The design principle of the robot should try to ensure that the personal privacy of the target person and make them are protected to the utmost extent.

\subsection{Starting points and key points}

\subsubsection{Separation and combination system design ideas}
When designing a relatively large technical solution or research solution, we usually do not consider the entire technical solution as an inseparable whole, from beginning to end, design step by step. Such a design strategy may make the whole design scheme very cumbersome in the revision and improvement of the later project, and may cause the design scheme to be disconnected, and the implementation strategy of such a design strategy is not very efficient too. Therefore, we adopt the idea of separation design to improve project design efficiency. When the research of each part of the system is completed, each part is combined to carry out the overall analysis and optimization of the project.

We divided the entire technical solution into 8 major studies, namely:
1. Mechanical and circuit structure design;
2. Sensor system construction;
3. Structure of the detection model;
4. Robot following and control model;
5. Introduction of gait analysis;
6. Filter analysis;
7. Exception handling model;
8. Functional and technical expansion.

Due to the time limit for writing this report, this project only addresses the first four parts of the above sections. The last four parts will be arranged in the second edition of the project research report for detailed explanation, so stay tuned.

\subsubsection{Separation and combination system design ideas}

For this issue, we provided the following idea: 1. Considering the actual use, the horizontal interface volume area of the robot should be similar to or slightly larger than the target person's volume. 2. When the robot is working at a close distance, the distance between the robot and the target person should not be too large. 3. The program should try to ensure that the robot is directly in front of the target person to achieve best experimental results. 4. When designing a robot, it should be fully realized that the final robot is to be put into practical use. Therefore, first of all, when designing the technical solution, we should fully consider the error of the detected data and the signal noise as well as the influence of specific environmental factors. For this practical application of technical solutions, the best way to assess their strengths and weaknesses is field experiment verification. Through the analysis of specific experimental results, efforts to optimize and upgrade the improved technical solutions. 5. The postures of human beings' walking are almost regular and habitual, and the walking gait of a person can reflect personal characteristics. It can even reflect some physiological and pathological features of the individual. Related to this is the field of gait analysis and gait recognition research. Therefore, we can take gait analysis into account when modeling human walking models.

\subsection{Detection error and prediction error}

In order to better analyze the errors caused by predicting the future path of humans, we divide the errors caused by predicting future human paths into two types: detection error and prediction error. The detection error refers to the error between the detected data and the real data when the human body parts are detected in real time within a fixed time. The prediction error refers to the error between the predicted human future path and the real human future path when the future path of the human is predicted in real time within a fixed time. Thinking from the definitions above, these two kinds of errors are relative. From the definitions we can find that the detection error measures the error in a short time, and the estimated error is the error measured over a long period of time. These differences are especially important when we set the size of the path that predicts the future path of humans. A direct reason is that the larger the distance of the predicted future path of the human being, the greater the initiative of the robot, because the robot can be shorter after knowing the predicted data information of the future path that is predicted to be sufficient for humans. A more complete response (e.g., stay in front of humans, avoid big turns, and automatically return to the front of humans after a turn). Conversely, if the predicted distance of the human future path is smaller, the initiative of the robot will be smaller, and the time left for the robot to react will be shorter.

But this does not mean that the larger the path to predicting the future path of mankind, the better. Combined with the definition of detection error and prediction error, if the predicted future path of human is shorter, the corresponding detection error will be larger, and the corresponding prediction error will be smaller. If the predicted distance of the human future path is longer, the corresponding detection error will be smaller, and the corresponding prediction error will be larger. This is because, the shorter the distance we set for the future path to be predicted, the less test data we need to analyze, and the fewer number of detection required. The corresponding detection data contains the detection error. Big. Because when the amount of detected data is relatively large, we can balance or reduce the corresponding detection error by means of data processing. Accordingly, if amount of detected data is smaller, the uncertainty of corresponding detected data is larger.

In general, we can only predict the future path of a short distance, that is, the predicted future path cannot be too large. Because the walking path and direction of travel of a human being are changeable, a walking human being may change his or her destination or direction of travel at any time, and this generally does not tell the robot who actively follows him or her. We cannot expect to predict a large future path of humans by detecting a small segment of human path, because errors.

However, we can't set the path of human prediction of the future path very short for the reason that to reduce the prediction error, because this will lead to an increase in the corresponding detection error. Since the robot does not know the detection error at this moment (n.b., this moment is meaning the real time in this report) during the execution of the follow-up process, although the robot can compare and predict the prediction error of the previous moment by comparing the predicted data with the predicted data. However, please note that the prediction error of the previous moment has no practical help for the prediction of the future path of the human being at the next moment. The detection error at this moment is only helpful for correcting the future path prediction of human beings at this moment. However, since the robot cannot obtain the detection error at this moment by itself when detecting at this moment, the robot needs to get the detection errors with external forces at this moment. For the reason that this report focuses on the existence of robots in the form of a single individual, the expansion of getting the detection errors with external forces will be explored through expanding discussion sections.

The main purpose of why we set the detection error and prediction error is to measure the accuracy of the prediction results. We use known data to predict unknown data, which means that we predict human future path data from human path data that has been detected. In general, the more known data is used, the more or the more accurate the future data can be predicted. However, due to the variability and temperament of the human walking path, this rule does not apply well in reality. In our proposed prediction model, human path data or gait data is used to predict future path data or future gait data of the human. For convenience of presentation, we will use the existing path data or gait data of humans to be predicted as sample data. We relate the detection error to the prediction error to the sample data. Therefore, the larger the data amount of the sample data selected at the present time, the larger the data amount of the path data of the future path of the human being predicted, and of course, the corresponding prediction error will become larger, but the corresponding detection error will be smaller. In other words, the amount of sample data must be appropriate, not too much or too little. The best solution is to find the data volume of the most suitable prediction data so that the prediction error and the detection error can reach a balanced state. That is, when predicting the future path of humans, the amount of sample data has an optimum value, so that the sum of the detection error and the prediction error is the smallest.

Promotion, we found through analogy and related experiments that all the phenomena that predict the future data through the detection data will have a relative tuning relationship between the detection error and the prediction error. Interested readers can further conduct relevant research.

\section{Multiple Methodology Analyses}

\subsection{Multiple pre-judgment}

In many current papers on Front-following technology researches, most of the detection methods are directly detect human location data and use them as predictive model data. There are also some papers that study the detection of specific parts of the human body to obtain test data (e.g., human legs or torso). But they are biased to detect only a certain part of the human body being tested. This can cause a lot of error in some extreme cases. Different from them, our detection model proposed by the technical solution will simultaneously detect multiple parts of the human body, and one of the advantages of this is that the accuracy of the prediction result can be effectively improved, and the error caused by some extreme situations can be well avoided. The specific details will be explained later.

\subsection{Gait prediction and gait advance}

The concept of path prediction has been mentioned in related researches introduced in many papers. In the Front-following techniques, the robot must predict the future path of the target person in real time in order to respond accordingly. In these studies, the researchers focused on calculating the most probable path of the target person in the future by a good algorithm, and calculating the best tracking execution path based on this path, and then driving the robot to track the motion. These studies focus on detecting the future path of the entire segment rather than dividing it into a number of small segments. One of the reasons for this is that studying a large whole path is enough to balance the error caused by studying only a small path. Therefore, the size of this small segment is worth thinking about. That is what the optimal size of the known walking path used to predict the target person should be. For ease of explanation, we will refer to the short path used to predict the future path of the target person as the sample path. From the concept of detection error and sample error, we know that the sample path is composed of sample data. So, again, the size of the sample path has the most appropriate amount. But what is discussed here is not the prediction of the future path of the target person, but the prediction of the future gait of the target person, which we call gait prediction. The shift in research objectives from sample data to subdivided sample data is the shift from the prediction path to the predicted gait. Obviously, the gait prediction research is to predict the gait action that the target person will take in the next short period of time for a certain moment gait of the target person.

For the reason that we can only predict future gait data information by gait data information that has been detected, gait prediction also has a relative relationship between detection error and prediction error. But in fact, the time for humans to take a normal step is usually very short. And under normal circumstances, human gait movements are relatively stable within the action range of human taking a step. Rarely occurs that human gait movements change dramatically within the action range of human taking a step. This is related to human walking habits and gait habits. Therefore, human gait prediction has high predictive directionality and guiding significance. We will add gait prediction to the predictive model, and only predict the possible action of the human gait in the range of motion of the step, that is, and predict the change of the future gait of the target person's leg from the time of lifting to landing.

Gait presupposes means that in a time when a certain human leg has been lifted but has not yet landed, the gait prediction mechanism predicts the future gait of the human next moment, and the robot adjusts the gait in real time and performs corresponding movements. It also in order to achieve the process of staying ahead of human gait in the next step. Therefore, the gait is a highly real-time following approach, which is more than enough to keep the robot in front of humans for follow-up without producing significant motion shifts. Our defined path prediction and path following are macroscopic prediction and following, while gait prediction and gait leading are microscopic. In specific predictive modeling, we will combine macro and micro predictions. In the specific control modeling, we will also combine the macro leading and micro leading. This is also based on the design concepts and principles of multiple pre-judgment. See the Predictive Models and Control Models section for more details.

\subsection{Pre-judgment restrictions based on ergonomics}

We know that the physical and skeletal structure of the human body limits the movement of human limbs and other parts of the body to a certain extent. Human walking involves multiple bones and muscle movements in humans. This makes the body's body trunk and limb movement have certain restrictions. Under this constraint law, we can predict some human behaviors and incorporate them into the prediction model to enhance the accuracy of the prediction model. For example, when a human raises his leg, the probability of the leg moving forward is much greater than the probability of backwards. This kind of knowledge can be applied to gait prediction. The idea of restricting the pre-judgment is that the action pre-judgment is carried out under certain restriction rules, so that the accuracy of the pre-judgment will be greatly improved. The restricted structure of anatomy can help us optimize the predictive model. See the Predictive Models and Control Models section for more details.

\section{Structure and Mechanism Analyses}

\subsection{Mechanical structure and 
circuit design}

Mechanical structure is an important part of a robot. The mechanical structure is equivalent to the skeleton of the robot, and the circuit is equivalent to the muscles and blood vessels of the robot. Robots combined with mechanical structures and electronic circuits already have a hardware foundation. In addition, robots need a soul. This soul is the algorithm. The algorithm belongs to the software part. Only the combination of mechanics, circuits and algorithms, the robot has the basis to achieve basic functions. Since this report focuses on the algorithmic implementation principle of the robot, the mechanical and circuit design will be discussed in detail in future upgrades, and only a brief description will be given.

The robot designed in this report is based on four-wheel drive. The premise of the implementation of the robot algorithm should be that the mechanical and electronic parts can respond well to the requirements of the software algorithm. Some of the mechanical and electronic designs will be discussed in the next section on sensor systems.

\subsection{Construction of sensor system}
To build a well-functioning sensor system, it is necessary to fully understand and analyze the functions and implementation conditions of various sensor candidates. Based on the requirements of our research project, we can use the following sensors to build the sensor system of the robot.

\subsubsection{Laser ranging sensors}
Laser sensors are mainly used to detect obstacles and interference from the humans realistic application scenarios and the environment. This kind of sensor is widely used in the design of robot following system, and it is also the most accurate range measuring sensor with mature technology. Widely used in ranging and positioning.

\subsubsection{Thermal infrared imaging sensors}
The biggest function of the thermal infrared imaging sensor is to be able to detect the thermal infrared rays emitted by the organism and to form a thermal infrared image. Since the personal privacy information about the target person displayed by the thermal infrared image is very small, it is feasible in ensuring personal privacy of the human being. Although the current thermal infrared technology may have a problem of insufficient accuracy, the thermal infrared imaging sensor is still an excellent sensor to choose from without precise positioning. We recommend the use of thermal infrared imaging sensors in this project to distinguish organisms from non-living organisms. Further, according to the detection area and temperature characteristics, humans are distinguished from other organisms, even under the premise of high progress. The target person is distinguished from other unrelated humans.

\subsubsection{Ultrasonic sensors}
The most prominent advantage of ultrasonic sensors is their low cost, and correspondingly, the cost is not high enough. Can be used as an alternative sensor solution. Ultrasonic sensors are optional in this project. The ultrasonic sensor is not described in the sensor system construction section, but will be described in the section on the detection and acquisition of the upgraded version report. 

Based on the design idea of separation and combination system and the design idea of multiple pre-judgment, we will scan the target person by scanning the whole body. Our detection method is not only for a certain part of the body of the target person, but for the whole body of the target person. In the whole body detection, based on the separation and combination system design idea, we divide the whole body of the target person into four main parts for detection: head detection, torso detection, legs detection, and feel detection. When detecting the body structure of the target person, the sensor we use is LRF, for the reason that to ensure the accuracy and reliability of the test data.

\subsection{Design of Sensor System}
We propose two sensor model system models, one based on a rectangular four-vertex angle model and one based on a rectangular four-sided center point model. Both modes have their own advantages and disadvantages. The model based on the rectangular four-vertice angle design has high detection precision for the detection areas at four corners, can perform double detection of a wide range, and can perform detection data superposition processing to achieve less detection error. The disadvantage of this model is that the installation structure is more complicated. The model design based on the rectangular four-sided center has the advantages of simple installation, stable scanning structure, strong front scanning capability, dual-mode data processing, and the ability to detect and track the influencing factors of the surrounding environment while the robot following target person. This scanning mode also has the advantage of performing a wide range of double detection data superposition to achieve less error. Moreover, the scanning detection accuracy of the scanning area in the four corners is also high, but not higher than the rectangular four-corner model. Due to the limitation of writing time, the project report will first discuss and compare the differences between the two scanning modes in the later algorithm control part. Then, in the following algorithms, all of them will be explained and analyzed based on the rectangular four-sided central model. And the model is very suitable for the dual-mode scanning algorithm proposed in this report. For other detailed analysis based on the rectangular four-corner model, please note the new version of the project design report.

\section{Conclusion of Sensor System}
For content arrangement of this paper series, we make the first simple conclusion in this secession. In this paper, we analyze the basic methods and principle background of a robot system with the sensor system, for more detailed technologies analyzes, note the other papers of this paper series. Sensor system design is significant for robot detection system design, for example, combine with the area drawing detection technology \cite{lin2022multi}, can improve the detection function of robot detection system.

% >>:=======================================================================================
\bibliographystyle{IEEEtran}
\bibliography{ref}{}
\end{document}